\documentclass{article}

\usepackage[english]{babel}

\usepackage[letterpaper,top=2cm,bottom=2cm,left=3cm,right=3cm,marginparwidth=1.75cm]{geometry}

\usepackage{amsmath}
\usepackage{graphicx}
\usepackage[colorlinks=true, allcolors=blue]{hyperref}

\title{ Conditioning GAN Without Training Dataset}
\author{Kidist Amde Mekonnen\\ (kidistamde.mekonnen@studenti.unitn.it)\\University of Trento}

\begin{document}
\maketitle

\section{Introduction}

Deep learning algorithms have a large number of trainable parameters often with sizes of hundreds of thousands or more. Training this algorithm requires a large amount of training data and generating a sufficiently large dataset for these algorithms is costly\cite{noguchi2019image}. 

GANs are generative neural networks that use two deep learning networks that are competing with each other. The networks are generator and discriminator networks. The generator tries to generate realistic images which resemble the actual training dataset by approximating the training data distribution and the discriminator is trained to classify images as real or fake(generated)\cite{goodfellow2016nips}. Training these GAN algorithms also requires a large amount of training dataset\cite{noguchi2019image}. 

In this study, the aim is to address the question, "Given an unconditioned pretrained generator network and a pretrained classifier, is it feasible to develop a conditioned generator without relying on any training dataset?"

The paper begins with a general introduction to the problem. The subsequent sections are structured as follows: Section 2 provides background information on the problem. Section 3 reviews relevant literature on the topic. Section 4 outlines the methodology employed in this study. Section 5 presents the experimental results. Section 6 discusses the findings and proposes potential future research directions. Finally, Section 7 offers concluding remarks.

The implementation can be accessed \href{https://github.com/kidist-amde/BigGAN-PyTorch}{here}.

\section{Background}

One of the methods used to overcome the shortage of datasets in the deep learning community is data augmentation. The data augmentation method is often used in machine learning to create synthetic datasets by altering the original data in a label-preserving way (that doesn’t change the semantics of the original data). In image processing rotation, reflection, shear, zoom, and shift transformations are used to create synthetic data from the original dataset. Often the synthetic images are generated on the fly while training the models. The data augmentation method has proved to improve the accuracy of the image classification algorithms by reducing the risk of overfitting \cite{krizhevsky2017imagenet}. 

For training GANs Image augmentation is used to overcome the overfitting of the discriminator. Zhao et.al \cite{zhao2020differentiable} applied differentiable augmentation to both the real and generated images. Karras et. al\cite{karras2020training}  used the Adaptive Discriminator Augmentation method(AD) to apply augmentation to both real and generated images.  The augmentation strength is chosen based on the discriminator performance on training, validation, and generated images. In order to deal with the overfitting of the Discriminator model, Sushko et. al\cite{sushko2024generating} used two types of discriminator networks, namely Content Discriminator and Layout discriminator. Both of the models see only partial information of the images. In addition, the authors also applied augmentation to both content and layout of the image.

Morerio et al.'s work \cite{morerio2020generative}  is closely related to this study. Their objective is to train a conditional GAN (cGAN) on a noise target distribution. First, the classifier is trained on source distribution with correct labels. Then the classifier is used to generate pseudo-labels on the target distribution. These pseudo labels are used to condition the cGAN. After some pretraining steps of the cGAN, the classifier is retrained jointly with cGAN on the generated images. The authors have argued that the shifting noise will not affect the quality of the generated images. Unlike the approach by Morerio et al., this method does not use a training dataset and does not involve training the generator network or the classifier network.

This study involves training additional neural networks to generate inputs for the GAN model. These networks accept one-hot encoded inputs corresponding to the attribute of interest. Although the method is task-agnostic, its performance is demonstrated on ethnicity classification. Additionally, the approach can be applied to a wide variety of GAN architectures and classifiers.

\section{Related work}

BigGANs are generative models which are trained by using class conditioning and scaling up a batch size and trainable parameters\cite{brock2018large}. The authors also used a truncation trick to control the trade-off between fidelity and sample variance of the generated images. BigGANs have been trained utilizing images from the UTKFace dataset \cite{parkhi2015deep} to generate facial images.

An ethnicity classifier has been trained using the UTKFace dataset. Pretrained VGG-Face models \cite{parkhi2015deep} were fine-tuned for this purpose. While experimenting with finetuning the ResNet model trained on the ImageNet dataset, it was observed that the model fine-tuned from the VGG-Face model exhibited better convergence speed and performance on the validation set. This performance difference is attributed to the domain variance between the original training dataset and the dataset used for fine-tuning.

BigGANs accept random inputs generated from a normal distribution. In this approach, the normal distribution used is generated by the input generator model. Simply generating inputs from a normal distribution with a mean of $\mu = 0$ and variance of $\sigma = 1 $ would make it impossible to propagate gradients to the input generator network. Therefore, a reparameterization trick \cite{kingma2013auto} is employed to address this issue.

\section{Method}
This study demonstrates the feasibility of creating a conditioned generator without requiring any training dataset, leveraging an unconditioned pretrained generator network and a pretrained classifier. Instead of directly modifying the GAN architecture to incorporate conditioning for image generation, an additional network is introduced to generate inputs for the GANs. The overall architecture of the model is depicted in Figure 1.

\begin{figure}[htb]
\begin{minipage}[b]{1.0\linewidth}
  \centering
  \centerline{\includegraphics[width=16cm]{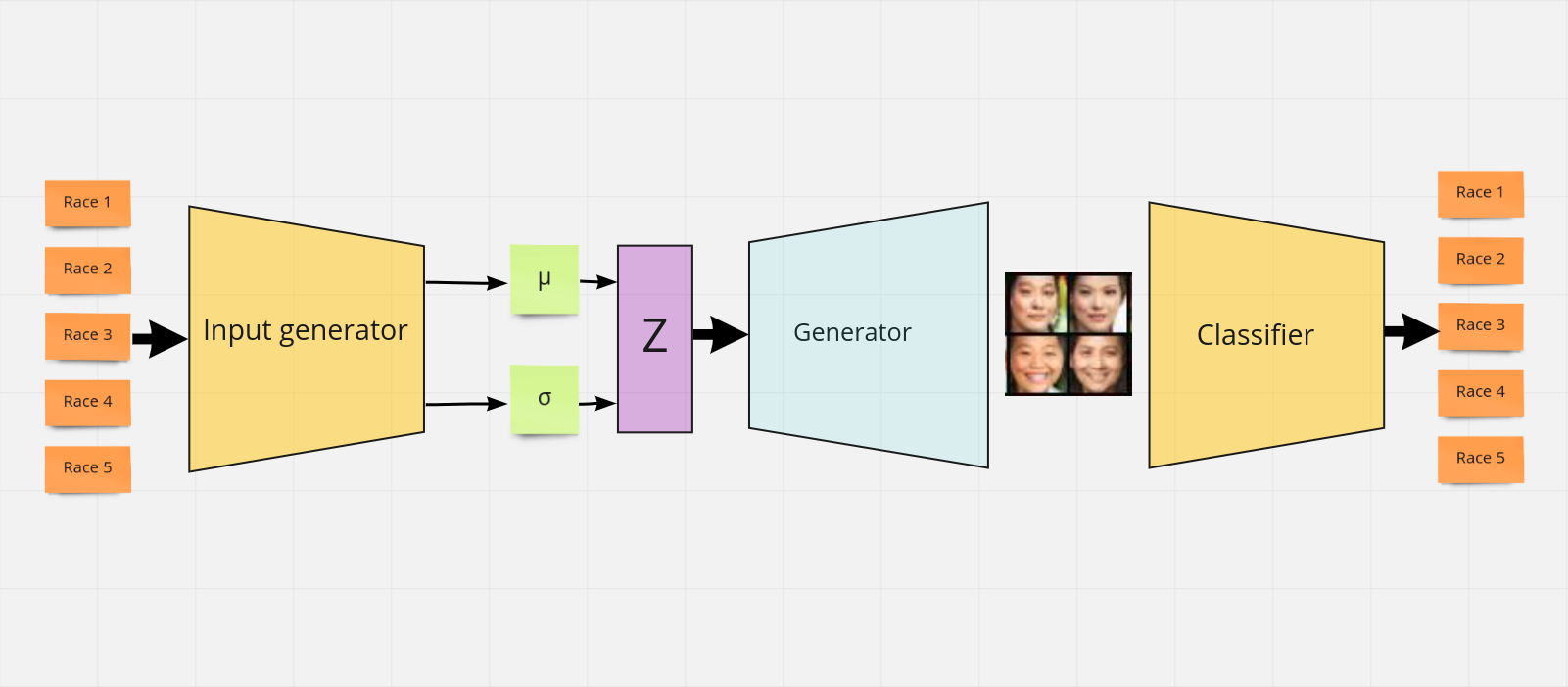}}
\end{minipage}
    \caption{Conditional GAN}
\label{fig:arc}
\end{figure}

The model comprises three primary components: the Input Generator, the Generator, and the Classifier.\\
\textbf{Generator}
The network under consideration is a pretrained generator of a GAN model. In this experiment, the generator network from the BigGAN model pretrained on the UTKFace dataset is employed. The weights of the generator network remain fixed. \\
\textbf{Classifier}
The ethnicity classifier deep learning model pretrained on the UTKFace dataset was utilized. Similar to the generator network, the classifier network’s weights are also fixed after pretraining.  Since the output space of the BigGAN(which has a size of 32x32x3) is different from the input space of the classifier(which has a size of 224x224x3)the nearest upsampling method was employed to scale up the images generated by the generator network.\\
\textbf{Input Generator}
This network is used to generate the inputs for the Generator network. In order to propagate the gradients from the loss through both classifier and generator to the input generator parameters, a reparameterization trick has been used. The input generator network generates two outputs for mean $\mu $ and variance $\sigma$. The input generator function can be written as:-

\begin{align}
   \mu = f_{\theta{1}}(x)\\
\sigma = f_{\theta{2}}(x) 
\end{align}

Where $\theta$1 and  $\theta$2 are the parameters of the input generator model. The subset of the parameters are shared between the $\theta$1 and $\theta$2. The input $x$ is one-hot encoded data of the ethnicity to be generated. Using the two outputs the input to the generator can be generated by sampling $z$ from a normal distribution with mean $\mu$ and variance $\sigma$. 
\begin{align}
   z \sim \mathcal{N} (\mu , \sigma^2)
\end{align}
With the reparameterization trick, this can be rewritten as
\begin{align}
   z = \mu + \sigma \epsilon
\end{align}
Where $ \epsilon \sim \mathcal{N} (0 , 1) $

Let’s denote the mapping of the generator as $f_{\theta{g}}(x)$ and mapping of the classifier with $f_{\theta{c}}(x)$. Then the output from the model for input $x$ can be written as:-
\begin{align}
z = f_{\theta{1}}(x) +  f_{\theta{2}}(x) \epsilon\\
i = f_{\theta_g}(z)\\
\hat{y} = f_{\theta_c}(i)
\end{align}

Where $z$ is the latent input to the generator, $i$ is an image generated by the generator network and \^{y}  is the output from the classifier. Then the loss can be computed as:-
\begin{align}
\mathcal{L}(\theta_1,\theta_2,x) = CCE(x,\hat{y})
\end{align}

Where the CCE is a cross-entropy loss function. From the above equation, it is clear that we can generate random one hot encoded input data and we can train the input generator by minimizing the cross-entropy loss between the classifier’s prediction and the input.

\section{Experimental Results}

BigGAN was trained with two different setups. In the first setup, the model was trained using the entire training dataset of UTKFace, while in the second setup, only 20\% of the training dataset was utilized. Table \ref{tab:fid} presents the performance of the models from both setups.

\begin{table}
\centering
\begin{tabular}{l|r|r}
Setup & Inception score & FID \\\hline
Full training data & 1.986 +/- 0.009 & 8.4984 \\
20 \% of training dataset & 2.111 +/- 0.021 & 45.2632
\end{tabular}
\caption{\label{tab:fid}Performance of the models}
\end{table}

As evident from Table \ref{tab:fid}, the model's performance is influenced by the size of the training set. However, the difference is minimal. Subsequent images illustrate the disparity between the images generated by the two setups. The generator trained on the full training dataset is employed for the method.

The ethnicity classifier was trained by fine-tuning the pretrained VGG-Face model, achieving a validation accuracy of 86\%. Initially, feeding the images directly from the generator to the classifier did not result in effective conditioning. Therefore, an upsampling algorithm was introduced between the networks to resize the generated images to a size of 224x224. After training for a few epochs, the model attained 100\% accuracy.

Table 2 displays the performance of the model on randomly sampled test data.

\begin{table}[ht]
\centering
\begin{tabular}{l|r|r|r|r}
 &Precision & Recall & F1-Score & Support \\\hline
European & 1.00 & 1.00 & 1.00 & 12721 \\
African & 1.00 & 1.00 & 1.00 & 12780\\
East Asian & 1.00 & 1.00 & 1.00 & 12869\\
South Asian & 1.00 & 1.00 & 1.00 & 12799\\
Other Ethnicities & 1.00 & 1.00 & 1.00 & 12831\\
Accuracy & & & 1.00 & 64000\\
Macro avg & 1.00 & 1.00 & 1.00 & 64000\\
Weighted avg & 1.00 & 1.00 & 1.00 & 64000\\
\end{tabular}
\caption{\label{tab:widgets}Performance of the models}
\end{table}

\section{Discussion and future work}
The model quickly learns to generate latent inputs for the specified features(ethnicity in our case). The performance of the model reaches 100\% on the validation set in as few as 8 epochs. The main reason for this case is the inputs generated by the input generator lack much variance. This has caused the outputs for certain classes to be around specific regions. Even though the variance of the image is not large, the model still generates a variety of face images because of the small noise added to the input generator.

\begin{figure}[htb]
\begin{minipage}[b]{1.0\linewidth}
  \centering
  \centerline{\includegraphics[width=8cm]{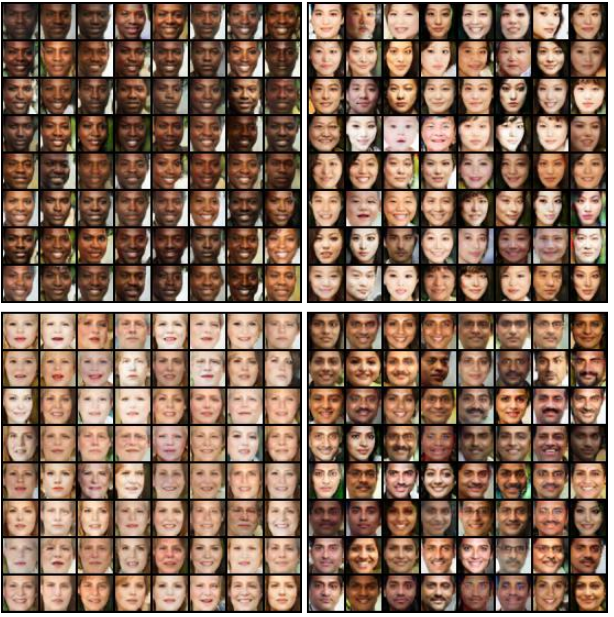}}
\end{minipage}
    \caption{Images generated by conditioning the GAN. Top-left: Conditioned with African ethnicity. Top-right: Conditioned by East Asian ethnicity. Bottom-left: Conditioned by European ethnicity. Bottom-right: Conditioned by South Asian ethnicity.}
\label{fig:images}
\end{figure}

The variance of the images generated by the model can be further increased by adding a regularization term that maximizes the variance. 

\section{Conclusion}

Training a GAN model typically necessitates a substantial training dataset to achieve better fidelity and diversity in generated samples. Moreover, conditioning the GAN further demands a large labeled training dataset. This study introduces a method to train a conditional GAN model utilizing pretrained GAN and pretrained classifier networks. Importantly, this method obviates the need for any dataset as long as pretrained generators and classifiers are available.

\bibliographystyle{IEEEbib}
\bibliography{strings,refs}
\end{document}